\documentclass[a4paper,twoside]{article}

\usepackage{epsfig}
\usepackage{subcaption}
\usepackage{calc}
\usepackage{amssymb}
\usepackage{amstext}
\usepackage{amsmath}
\usepackage{amsthm}
\usepackage{multicol}
\usepackage{pslatex}
\usepackage{apalike}
\usepackage[bottom]{footmisc}
\usepackage{xspace}

\usepackage{algorithm}%
\usepackage{algorithmicx}%
\usepackage{algpseudocode}%

\newcommand\Algphase[1]{%
	\vspace*{-2ex}\Statex\hspace*{-2.5ex}\rule{\columnwidth}{0.4pt}%
	\vspace*{-0.3ex}\Statex\hspace*{-2ex}\textbf{#1}%
	\vspace*{-2ex}\Statex\hspace*{-2.5ex}\rule{\columnwidth}{0.4pt}%
}
	
\makeatletter
\DeclareRobustCommand\onedot{\futurelet\@let@token\@onedot}
\def\@onedot{\ifx\@let@token.\else.\null\fi\xspace}

\def\eg{\emph{e.g}\onedot} 
\def\ie{\emph{i.e}\onedot} 
\def\cf{\emph{cf}\onedot} 
 \def\vs{\emph{vs}\onedot}

\def\etal{\emph{et al}\onedot}
\usepackage{url}
\usepackage{xcolor}%
\usepackage{booktabs}

\usepackage[capitalize,sort&compress,nameinlink]{cleveref}
\crefname{section}{Sec.}{Secs.}
\Crefname{section}{Section}{Sections}
\Crefname{table}{Table}{Tables}
\crefname{table}{Tab.}{Tabs.}

\usepackage{SCITEPRESS}     
\newcommand{\boldhline}{\specialrule{0.15em}{0em}{0.1em}}

\begin{document}

\title{Domain-Incremental Semantic Segmentation for	Autonomous Driving under Adverse Driving Conditions}

\author{\authorname{Shishir Muralidhara\sup{1,2}, René Schuster\sup{1,2} and Didier Stricker\sup{1, 2}}
\affiliation{\sup{1}German Research Center for Artificial Intelligence (DFKI), Trippstadter Straße 122, Kaiserslautern, Germany}
\affiliation{\sup{2}RPTU - University of Kaiserslautern-Landau, Gottlieb-Daimler-Straße 47, Kaiserslautern, Germany}
\email{\{shishir.muralidhara, rene.schuster, didier.stricker\}@dfki.de}
}

\keywords{Continual Learning, Continual Semantic Segmentation, Domain-Incremental Learning}

\abstract{
Semantic segmentation for autonomous driving is an even more challenging task when faced with adverse driving conditions. 
Standard models trained on data recorded under ideal conditions show a deteriorated performance in unfavorable weather or illumination conditions. 
Fine-tuning on the new task or condition would lead to overwriting the previously learned information resulting in catastrophic forgetting. 
Adapting to the new conditions through traditional domain adaption methods improves the performance on the target domain at the expense of the source domain. 
Addressing these issues, we propose an architecture-based domain-incremental learning approach called Progressive Semantic Segmentation (PSS). 
PSS is a task-agnostic, dynamically growing collection of domain-specific segmentation models. 
The task of inferring the domain and subsequently selecting the appropriate module for segmentation is carried out using a collection of convolutional autoencoders. 
We extensively evaluate our proposed approach using several datasets at varying levels of granularity in the categorization of adverse driving conditions. 
Furthermore, we demonstrate the generalization of the proposed approach to similar and unseen domains. 
}

\onecolumn \maketitle \normalsize \setcounter{footnote}{0} \vfill

\section{\uppercase{Introduction}}
\label{sec:introduction}

Autonomous driving systems perform well under ideal conditions, as they are typically trained using data captured under these conditions.
However in the real-world, data drift occurs, and the model is faced with adverse conditions such as weather and low illumination.
These factors tend to alter the characteristics and visibility of objects, causing a significant drop in the model performance.
Fine-tuning on the new distribution will result in overwriting of previously learned information, resulting in catastrophic forgetting \cite{catastrophic_forgetting}. 
This overwriting of information stems from the rigidity of neural networks.
Catastrophic forgetting can be circumvented with \textit{joint training}, where the model is trained with all the encountered data jointly, instead of learning sequentially.
However, this may not be possible due to data unavailability, storage constraints, computational and time costs of retraining the entire model with vast amount of data.
Conventionally, domain adaptation (DA) methods are used for adapting to data drift in the new or target domain. 
Domain adaptation often requires source data and focuses primarily on the performance on the target domain.
However, it is imperative for autonomous systems to constantly adapt to changing conditions, whilst maintaining performance across all domains.
\begin{figure}
	\centering
	\includegraphics[width=\columnwidth]{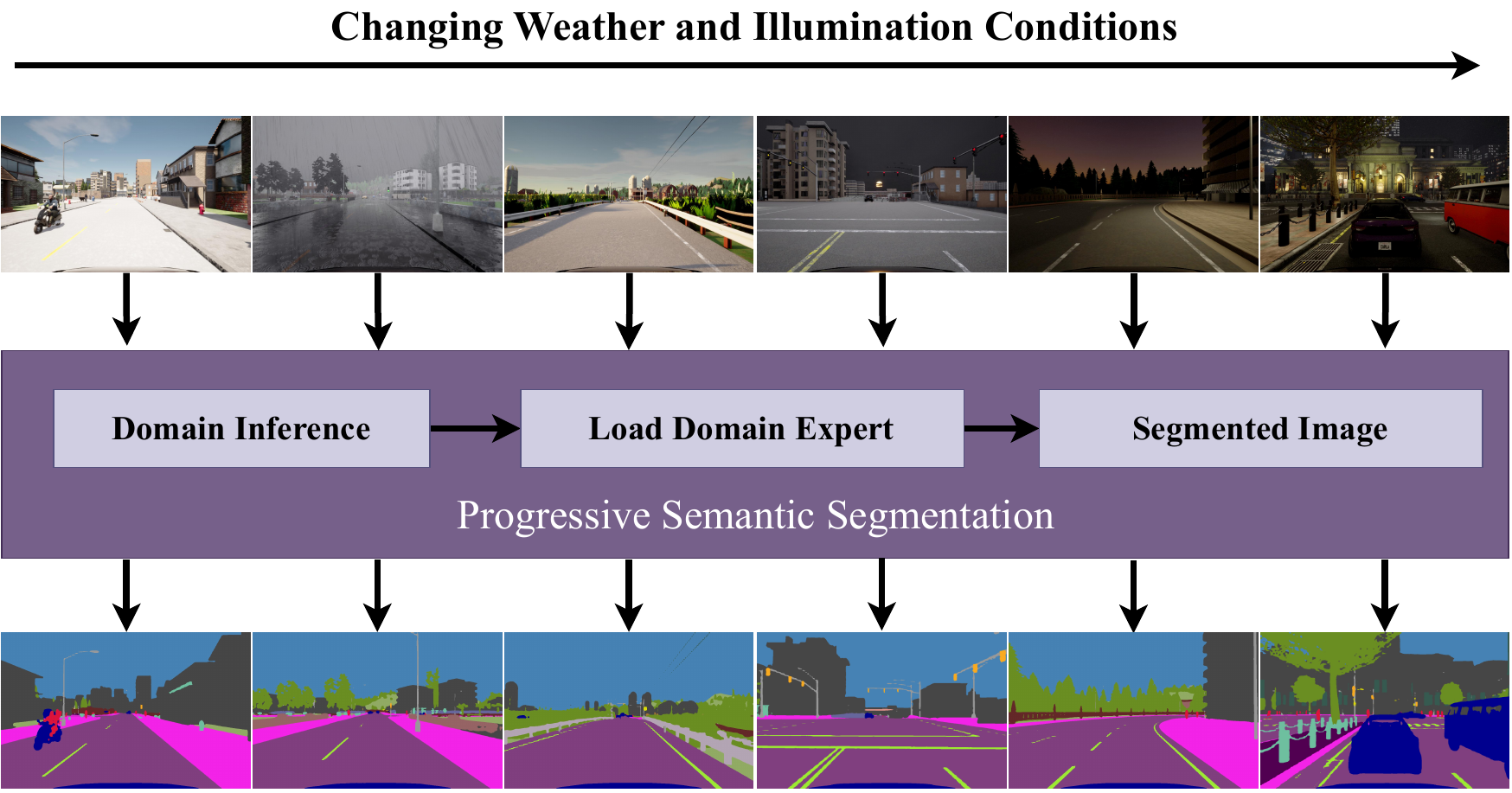}
	\caption{Progressive Semantic Segmentation (PSS) continually learns to handle adverse conditions. Our proposed approach accommodates to changing weather and illumination conditions by first inferring the domain and subsequently using a domain expert for segmentation.}
	\label{fig:teaser}
\end{figure}
Continual learning (CL) is a dynamic learning paradigm, that extends a trained model to the changing data and objectives, while addressing the above challenges.
CL is defined under the constraints of having no access to data from previous tasks. 
CL incrementally learns, and avoids the costs associated with retraining from scratch.
CL addresses catastrophic forgetting and emphasizes performance across all domains.
To do so, it tackles the stability-plasticity dilemma \cite{stability_plasticity}, a trade-off where the model must be able to learn new information on the current task, without forgetting the previously learned information.
In this work, we propose Progressive Semantic Segmentation (PSS), a CL based approach for continuous learning of adverse conditions as a problem of domain-incremental learning (\cf \cref{fig:teaser}).
The main contributions are outlined as follows:
\begin{itemize}
	\item PSS is a task-agnostic, architecture-based approach with a growing collection of domain-specific models as domain experts for segmentation under adverse conditions. 
	
	\item Unlike existing architecture-based DIL methods that require task-IDs, PSS  leverages autoencoders as task experts to infer the domain and select the most suitable model during inference.
	
	\item We validate the effectiveness of PSS using multiple datasets at varying levels of granularity in categorizing adverse driving conditions. 
	
	\item We extend our framework to other computer vision tasks such as object detection, including a novel hybrid incremental setting with both varying input and output distributions.
	
\end{itemize}

\section{\uppercase{Related Work}}

\label{section:background}

\subsection{Continual Learning} \label{section:bg:cl}
Continual learning methods can be broadly categorized into: Architecture-, regularization-, and replay-based methods. 
Architecture-based approaches incrementally learn by altering the network architecture.
The modifications can be implicit through adaptive and task-specific weights \cite{piggyback}, path routing \cite{pathnet} or explicit with dynamically growing progressive networks \cite{progressiveNN}.
Regularization approaches include penalty computing which prevents the model from overwriting parameters important to previous tasks \cite{EWC}, and knowledge distillation to transfer knowledge between tasks.
Rehearsal and replay are two closely related approaches, which use samples from previous tasks during training for the current task. 
Rehearsal based methods explicitly store a subset of previous task data, whereas replay uses generative models \cite{GANReplay} to sample instances.
Each approach is associated with advantages and limitations, and selecting an approach is dependent on the problem and available resources.
PSS is an architecture-based method that benefits from using domain-specific models to continuously learn different environmental conditions.

\subsection{Incremental Semantic Segmentation}
Incremental learning in CL can be formulated as three scenarios: In domain- (DIL) and class-incremental learning (CIL), the input (domain) or the output distribution (classes) is extended from task to task. 
While DIL and CIL are task-agnostic, in task-incremental learning (TIL), a task-ID is assumed to be known during inference.
MDIL \cite{MDIL} is an architecture-based approach with a shared encoder network comprising of universally shared and domain-specific parameters and domain-specific decoders.
It requires the task-ID during inference to select the domain-specific path. In PSS, we alleviate this requirement and dynamically infer the task-ID.
For image classification, the three types have been investigated comparatively deeply \cite{ExpertGate,DIL_Classifier} and research is moving forward to the more complex task of pixel-wise classification, \ie segmentation \cite{ILT,PLOP,goswami2023attribution}.
For class-incremental semantic segmentation, there exist diverse methods covering regularization-based approaches, \eg ILT \cite{ILT}, PLOP \cite{PLOP}, as well as replay-based RECALL \cite{RECALL}.
Kalb \etal \cite{CLScenarios} investigate the use of distillation and replay-based approaches for both DIL and CIL and observe that distillation is more suited for the former and replay-based for the latter.
Though PSS is designed for DIL, we demonstrate that PSS is also suitable for a combined incremental learning of new domains and new classes.

\begin{figure*}[t]
	\centering
	\includegraphics[width=\textwidth]{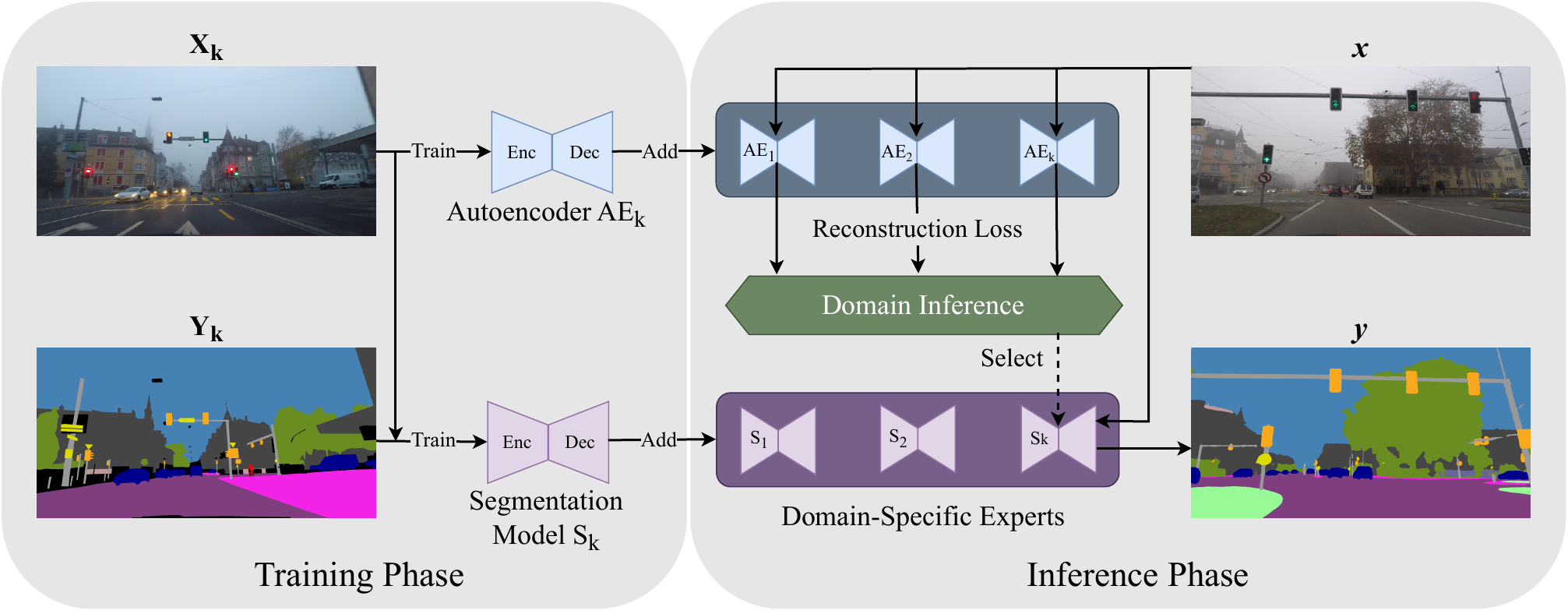}
	\caption{Overview of the proposed Progressive Semantic Segmentation (PSS). For each task-increment $T_k$ and the associated data $D_k=(X_k, Y_k)$, we train a task-specific autoencoder $AE_k$ using $X_k$ and a segmentation model $S_k$. During inference, the test image $x$ is reconstructed using autoencoders from all tasks, and the reconstruction losses are computed. The domain is inferred from the autoencoder with the lowest loss and the image is routed to the corresponding segmentation model.}
	\label{fig:Overview}
\end{figure*}

\subsection{Adapting to Adverse Conditions}

Domain adaptation (DA) methods emphasize the performance on a single target domain disregarding any previous source domains.
CL approaches strive to preserve the performance of the previous domains whilst adapting to the new domains. 
Additionally, DA relies on source and target domain data, contradicting the assumption in CL that data is available for a single task at a time.
However, strategies for DA can be used for DIL, if all subsequent domains are converted to mimic a common domain-specific condition, \eg by style transfer or light enhancement \cite{SFNet}. 
Style transfer can be used either during training with the converted dataset or during inference by converting the adversarial conditions into favorable conditions prior to segmentation \cite{BridgingGap}.
With PSS, we can avoid any intermediate transfer, which provides an additional source of error.
Several works \cite{NightTime,dannet} have proposed to tackle domain gaps through a sequence of smaller adaptations. 
Dark model adaptation \cite{NightTime} uses a model trained on daytime conditions to generate pseudo-labels for twilight images, which is used for training a model on nighttime images. 
DANNet \cite{dannet} additionally uses an image relighting network to minimize the intensity distributions between the domains. 
PSS is not a DA method, but a CL approach.

\section{\uppercase{Progressive Semantic Segmentation}}
\label{section:methods}

In a continual learning setting, the set of tasks $T$ arrives sequentially in increments $T_1, T_2,...,T_n$. Each task consists of a set of images $X$ and the corresponding pixel-level ground truth $Y$, with $C$ number of classes.
The increments between tasks can vary in terms of the input or output distribution \cite{CLScenarios}.
This work corresponds to DIL where new input distributions are added sequentially representing the changing adverse conditions.
The set of classes $C$ remains the same across all the domains.
Inspired by progressive neural networks \cite{progressiveNN}, our idea is to instantiate one domain-specific model per task.
However, to transfer this task-incremental setting into a task-agnostic one, we need to dynamically infer the domain.
We address this issue using a collection of autoencoders (AEs) similar to \cite{ExpertGate}.
We call the set of AEs \textit{task experts} and the set of segmentation models \textit{domain experts}.
An overview of our proposed approach (PSS) is presented in \cref{fig:Overview}.

\begin{algorithm}[t]
	\caption{Progressive Semantic Segmentation}\label{alg:pss}
	
	\textbf{Require:} Collection of task experts (TE) and domain experts (DE).
	\textbf{if} $k=0$ initialize TE and DE to [].
	
	\begin{algorithmic}
		
		\Algphase{Training Phase}
		\textbf{Input:} Task $T_k$ from set of incrementally added tasks $T_n$ and the associated data $D_k = (X_k, Y_k)$. 
		\State \textbf{Train} autoencoder $AE_k$ on $X_k$ and append to TE.
		\State \textbf{Train} segmentation model $S_k$ on $D_k$ and append to DE.
		
		\Algphase{Inference Phase}
		\textbf{Input:} Test image $x$ from unknown domain. 
		\newline Initialize reconstruction losses (RL) = []
		
		\For {each $AE_i$ in TE}
		\State {Reconstruct $x$ using $AE_i$, compute the reconstruction loss and append to RL}
		\EndFor
		
		\State \textbf{Domain Inference:} $domain = index(min(RL))$
		
		\State \textbf{Select Model:} $domainExpert = DE[domain]$
		
		\State \textbf{Segment Image:} $y = domainExpert(x)$
		
	\end{algorithmic}
\end{algorithm}

\cref{alg:pss} explains the training and inference sequence for a given task $T_k$.
For each task, we train an AE with the associated RGB images. 
During inference, the test image passes through each AE and the corresponding reconstruction loss is calculated.
The domain corresponding to the task-expert with the lowest reconstruction loss is inferred and the image is routed to the associated domain expert.
The set of very light-weight AEs can grow dynamically and allows for fast estimation of the domain.
Since each autoencoder is trained independently on individual tasks, the continual introduction of new tasks does not affect any of the previous models, thus typical challenges in CL do not affect the AEs.
This is a big advantage of the collection of AEs, compared to \eg a single domain classifier that needs to learn continuously.
Our approach of using small-scale AEs is scalable, avoids retraining, and does not require additional CL methods to mitigate forgetting.

\subsection{Autoencoders for Domain Inference}
\begin{figure}[t]
	\centering
	\includegraphics[width=\linewidth]{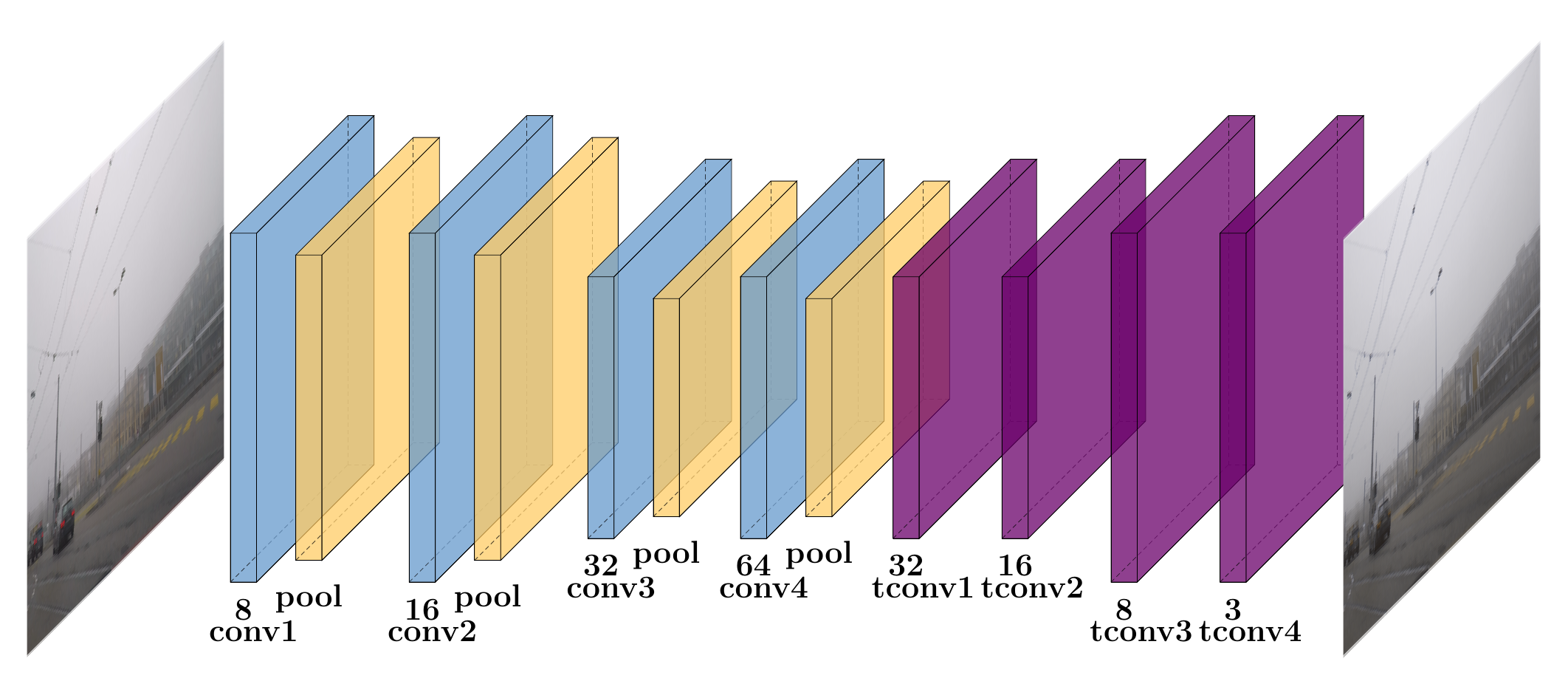}
	\caption{Proposed autoencoder architecture with four-layer deep encoder and decoder. 
		Domain inference is based on the difference between reconstructed and input image.}
	\label{fig:CAE}
\end{figure}

Autoencoders are primarily used for reconstruction tasks, where the goal is to reconstruct the input data from the compressed representation.
They consist of two components: An encoder network which maps the input data to a low-dimensional representation and a decoder network for mapping it back to the input space.
We use AEs to infer the domain during inference based on the reconstruction loss.
The reconstruction loss is a measure of the difference between the original input and the reconstructed output.
We use a simple convolutional AE with a four layers deep encoder and decoder as shown in \cref{fig:CAE}.
Our activation function is a ReLU and the final layer is activated by a Sigmoid function.
In the encoder, each convolution layer has a kernel size of 3 and applies padding, followed by a $2 \times 2$ max-pooling layer with stride 2. 
The decoder consists of transposed convolutions, exclusively.
Their kernel size is $2\times2$ with a stride of 2.
This model is very small, containing just 0.035M parameters, and has a size of $\sim142$ KiB. 

It is important to note, that the proposed PSS is not very sensitive to the design of the AE.
The architecture can almost be arbitrarily small.
Neither the size nor shape of the latent space matters.
Also, the quality of the reconstruction is not of primary interest, as long as the reconstruction loss can be reduced sufficiently.
Due to the rigidity of neural networks, any shift in the domain will result in a worse reconstruction compared to samples from the original distribution.
This builds the basis for a decision boundary when inferring the domain.

\subsection{Domain Experts}
Semantic segmentation involves assigning a semantic label to each pixel in the image and thereby segmenting an image into object regions.
In our work, we use DeepLabV3 \cite{DeepLabV3} based on an encoder-decoder architecture with atrous spatial pyramid pooling (ASPP) module.
The ASPP module with dilated convolutions leverages multi-scale context information.
A ResNet-101 \cite{ResNet} pre-trained on ImageNet \cite{ImageNet} is our backbone.
More specifically, we use the ResNetV1c variant of ResNet where the 7x7 conv in the input stem is replaced with three 3x3 convs.
Again, we highlight that PSS has no dependency on the specific segmentation model used.
The architecture of the domain experts can easily be replaced by \eg a more efficient or powerful network.
In fact, within our framework, the task itself can be replaced.
We demonstrate this in our experiments (see \cref{sec:results:detection}) by performing Progressive Object Detection (POD) under adverse conditions.

\section{\uppercase{Experiments and Results}}
\label{section:results}

In this section, we present the results of our approach in DIL of adverse conditions.
Progressive Semantic Segmentation (PSS) is primarily compared against three baselines:
The \textit{single-task} (ST) baseline, in which individual models are trained on each domain. The evaluation protocol for this baseline assumes availability of the task-ID similar to TIL.
The \textit{fine-tuning} (FT) baseline, \ie a single model is trained sequentially on the individual domains. The highest amount of forgetting is assumed in this scenario.
The \textit{joint training} (JT) model, which has been trained with all the data of all incremental steps at once. This model serves as a theoretical upper limit, as the availability of all data is restricted in CL.

The results are presented in terms of mean Intersection-over-Union (mIoU), calculated as the average of IoU values across all classes.
The IoU is the ratio of the area of overlap to the area of union between the predicted and ground truth segmentation.
In our experiments, we assess the amount of knowledge of a model compared to the single-task baseline.
The \textcolor{blue}{information gained} is highlighted in blue, \textcolor{red}{information lost} in red, and the \textcolor{gray}{information not learned}, because of too high stability, is highlighted in gray.
Our experiments cover a wide range of datasets, even some that have not been used for training.
Furthermore, we evaluate the capabilities of the AEs as domain classifiers and compare PSS to previous work.

\subsection{Datasets} \label{sec:results:data}
We evaluate our approach using several datasets of varying conditions and at different levels of granularity.
Some of the datasets are used exclusively for testing to highlight the generalization of the proposed approach to new, unseen domains as in a real-world setting.
The datasets used are described as follows.
\begin{itemize}
	\item \textbf{Cityscapes (CS)} \cite{CS} is a widely used autonomous driving dataset consisting of 2975 training and 500 validation images captured during ideal daytime conditions from different cities. It comprises 19 semantic classes.
	
	\item \textbf{Adverse Conditions Dataset with Correspondences (ACDC)} \cite{ACDC} consists of 1600 training and 406 validation images captured under conditions such as night, snow, rain, and fog. ACDC shares the label space of CS. 
	
	\item \textbf{SHIFT} \cite{SHIFT} is a large synthetic driving dataset consisting of 22 classes. We split the data into five non-overlapping categories of day, night (under clear conditions), rain, fog, and overcast (under daytime conditions).
	
	\item \textbf{Dark Zurich} \cite{DarkZurich} and \textbf{Nighttime Driving} \cite{NightTime} consists of images captured in the dark. We use the labeled test set for the evaluation of our approach in unseen domains. Both datasets consist of the same 19 classes of CS.
	
	\item \textbf{Indian Driving Dataset (IDD)} \cite{IDD} is recorded in less structured (crowded) environments and has a larger label space of 26 classes. As opposed to the label space of CS, IDD introduces several new classes while also making further distinctions in the classification of CS.
\end{itemize}

\subsection{Training} \label{sec:results:training}
The AEs are trained on a single GPU using a batch size of 8 and Adam optimizer with a learning rate of 0.001.
As reconstruction loss, we minimize the mean squared error (MSE) during training and we train each AE until the loss reaches below a satisfactory threshold, in our case 0.002.
We do not augment or pre-process the images in any way to best capture the nature of every domain.
For training of the segmentation model, we follow two training schemes in our experiments.
The first is the official implementation of PLOP \cite{PLOP} and is used for the experiments on CS and ACDC in \cref{sec:results:sota}.
The second is the more advanced pipeline of MMSegmentation \cite{mmseg}, used for all other experiments. 

\begin{figure}[t]
	\centering
	\includegraphics[width=\linewidth]{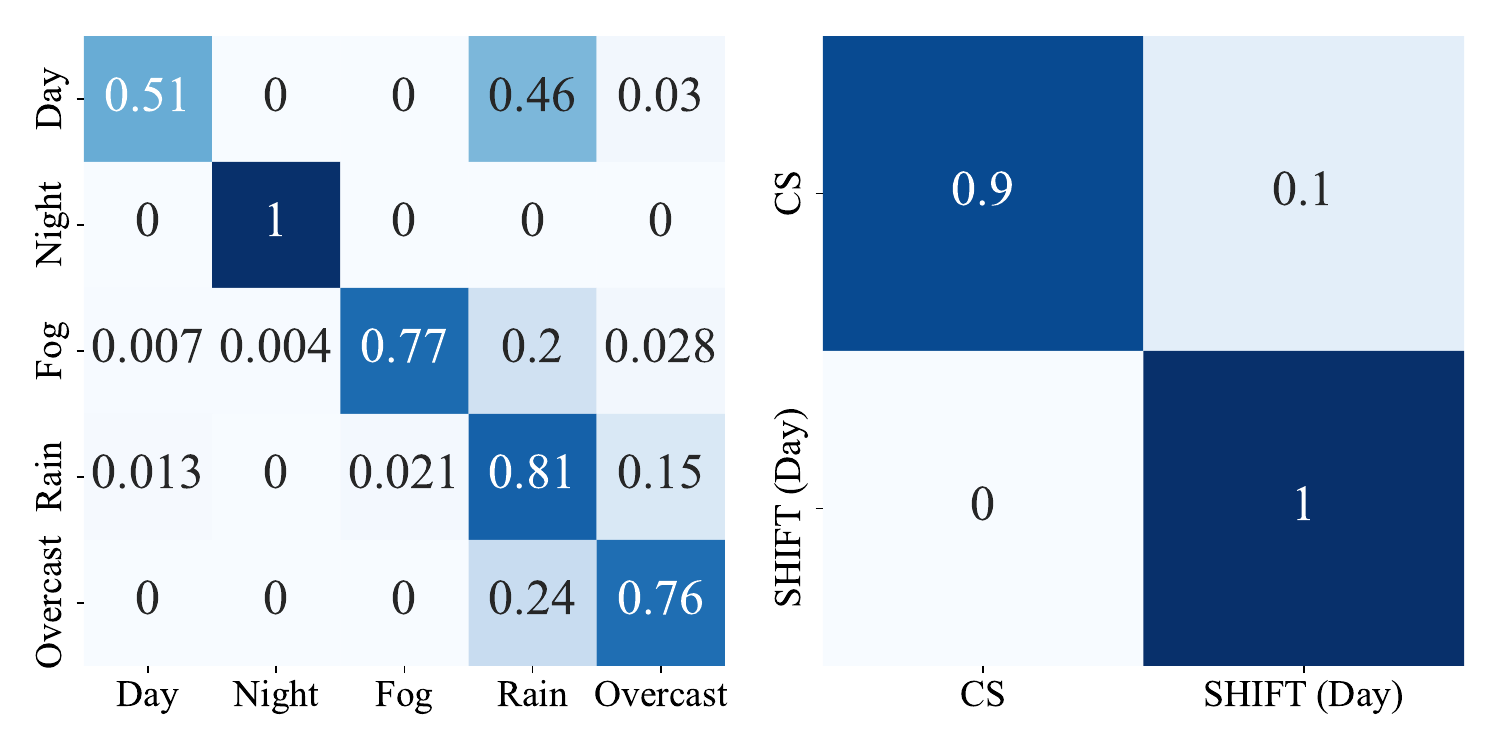}
	\caption{Classification results for domain inference.
		Left: SHIFT treats each adversarial condition as a separate domain, resulting in a multi-class classification.
		Right: Real vs synthetic data both representing daytime conditions, adds complexity to classification.}
	\label{fig:CFM}
\end{figure}

\subsection{Domain Inference} \label{sec:results:ae}
As discussed previously, we use the reconstruction by the task-experts as the basis for determining the domain.
This approach as opposed to using a domain classifier circumvents the need for further CL interventions when learning a new task.
We observe that when using a classifier for learning a single class representing the current domain, the model begins to overfit immediately.
When this overfitted model is subsequently trained on the next domain, it demonstrates a similar pattern of overfitting for that particular class and completely fails in predicting the previous class.
In contrast, our approach considers each domain independently and ensures there is no interference and overwriting of information from the previous tasks.
Additionally, standard classifiers tend to have significantly larger sizes compared to AEs, imposing additional memory constraints.
Interestingly, we achieve an accuracy of 100~\% for the classification between CS and ACDC, and therefore PSS will achieve results on par with the single-task baseline.
\newline
For SHIFT data, the multi-class classification is more challenging, yet we achieve an accuracy of 77~\% and the results are presented in \cref{fig:CFM}.
In \cref{sec:results:SHIFT} we show that, despite the lower accuracy, the segmentation results improve over the single-task baseline for a few domains.
This can be attributed to a selection of the most appropriate domain expert regardless of the true domain label.
To further affirm the effectiveness of domain inference with AEs, we conduct an experiment with real and synthetic data using CS and SHIFT, both representing daytime conditions. 
Despite their similarities, our approach accurately distinguishes between them, as shown in \cref{fig:CFM}, highlighting AE's capability to capture underlying features across domains, discerning subtle differences.

\subsection{Comparison on Real-World Data}
\label{sec:results:cs_acdc} \label{sec:results:sota}

\begin{table}[t]
	\setlength{\tabcolsep}{1pt}
	\centering
	\caption{Results on real-world data with a coarse distinction between ideal and adverse conditions. All models are first trained on the initial task (CS) and then on the adverse domain (ACDC). Our proposed approach alleviates forgetting completely and is close to the upper bound.}
	\label{table:comparison}

	\resizebox{\linewidth}{!}{
		\begin{tabular}
			{c@{\hspace{0.15cm}}c@{\hspace{0.05cm}}c@{\hspace{0.1cm}}c@{\hspace{0.05cm}}c@{\hspace{0.15cm}}c} 
			
			\boldhline
			\textbf{Method}  & \textbf{CS} & & \textbf{ACDC} &  & \textbf{Avg.}\\
			\boldhline
			
			Single Task & 61.53 &  & 59.53 & & --  \\
			\hline
			
			Fine-Tuning & 41.60 & \textcolor{red}{(-19.93)} & 61.80 & \textcolor{blue}{(+02.27)}& 51.70 \\
			\hline
			
			ILT & 37.62 & \textcolor{red}{(-23.91)} & 36.24 & \textcolor{gray}{(-23.29)} & 36.93\\

			Replay & 29.94 & \textcolor{red}{(-31.59)} & 59.65 & \textcolor{blue}{(+00.12)} & 44.79\\
			
			\textbf{PSS (Ours)} & 61.53 & \textcolor{blue}{(+00.00)} & 59.53 & \textcolor{blue}{(+00.00)} & \textbf{60.53} \\
			\hline
						
			Joint Training & 61.98 & \textcolor{blue}{(+00.45)} & 61.36 & \textcolor{blue}{(+01.83)} & 61.67 \\
			\boldhline
			
		\end{tabular}
		}
\end{table}

ACDC and CS are captured in real-world settings and together provide the basis for our first set of experiments.
We treat the entire ACDC dataset as a single class of adverse conditions (nighttime, snow, rain, and fog), while CS represents the ideal conditions. 
This results in a coarse distinction between ideal and adverse conditions.
We use this setting for most of our comparisons against other methods and some additional experiments.
The results are presented in \cref{table:comparison}.
We quantify the catastrophic forgetting associated with FT.
In case of JT, there is a minuscule improvement over the individual models, due to the more diverse training data.
With our approach, we observe that no knowledge is forgotten, and we achieve results of the single-task baseline as the AEs route all samples to their corresponding domain-specific expert with 100~\% accuracy.
ILT is a regularization-based method that freezes the encoder from the previous step and distillation is used to retain knowledge from the previously seen tasks.
The results in \cref{table:comparison} indicate that ILT alleviates catastrophic forgetting to a certain degree, but still, a significant amount of information is lost (too low stability, too high plasticity).
Training on the subsequent task subject to a distillation loss restricts the model from learning to the fullest, and it does not achieve satisfactory results (too low plasticity, too high stability).
With the increasing number of tasks, these two problems become amplified.

\begin{figure}[t]
	\centering
	\includegraphics[width=\linewidth]{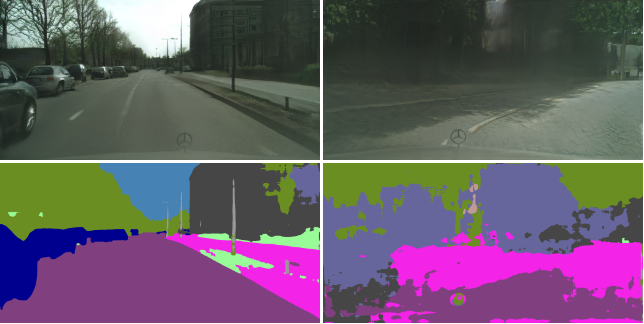}
	\caption{Examples of images for Cityscapes generated by GANformer \cite{ganformer} and the corresponding pseudo labels using the CS domain expert. The sample on the left is reasonably accurate, while the right sample seems unrealistic and provides erroneous labels.}
	\label{fig:CS_GAN}
\end{figure}

For the replay-based approach, we use GANformer \cite{ganformer}. We use the provided pre-trained model which generates high-resolution images of the size 2048x1024, and we generate 2975 images for training similar to the size of the original train set of CS.
Subsequently, we generate the corresponding pseudo-labels using the previous task model, \ie the domain expert for CS.
During training on the ACDC, we also replay the generated training samples.
Though the additional training samples help to obtain a positive forward transfer, the forgetting is even higher than fine-tuning.
This can be attributed to error propagation of the generated images and labels as indicated by the visualization in \cref{fig:CS_GAN}.

\begin{figure*}[t]    
	\centering
	\includegraphics[width=\textwidth]{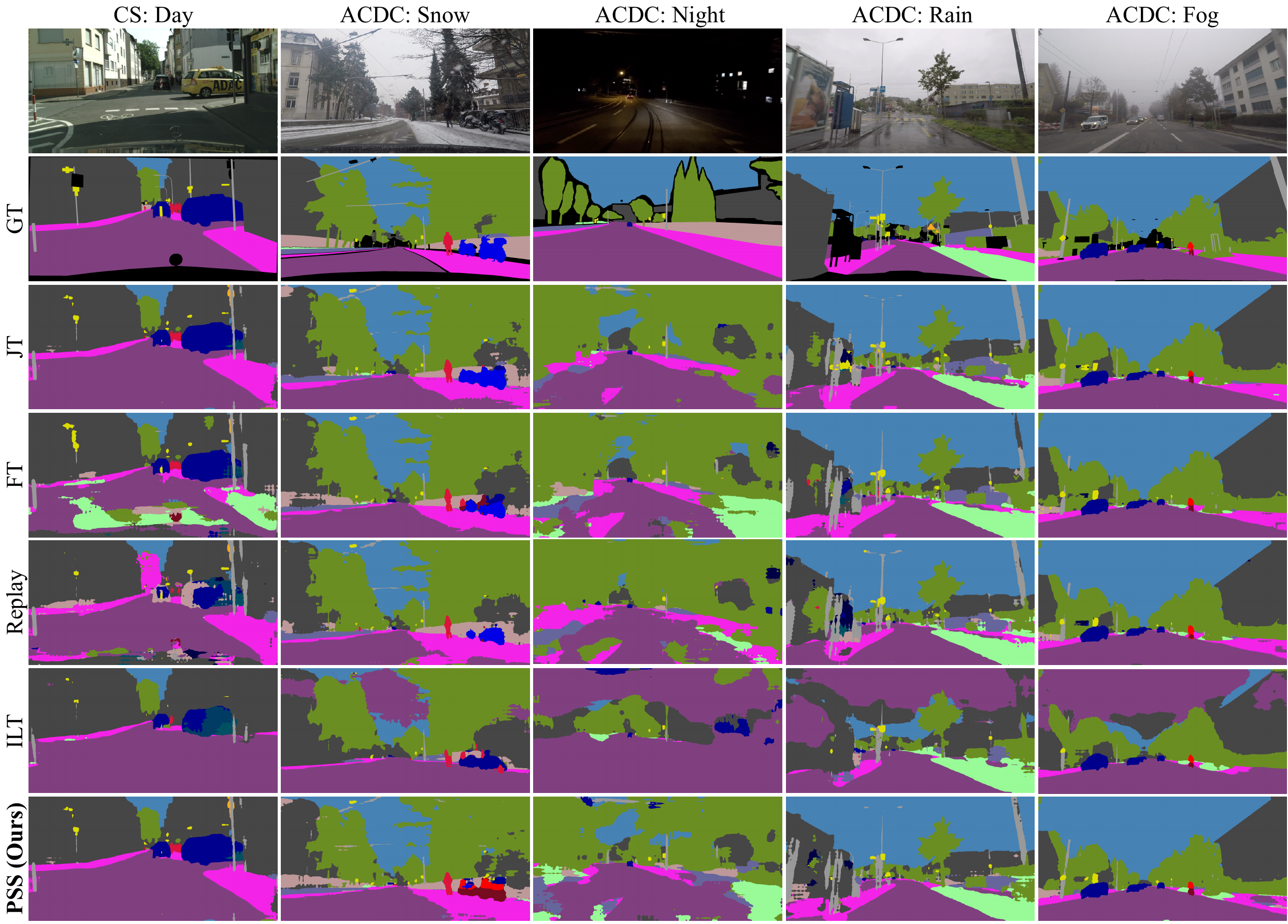}
	\caption{Qualitative visualization of predictions on CS and ACDC. Progressive Semantic Segmentation (PSS) achieves results of the corresponding single-task models on par with the joint-training (JT) which forms the upper bound.}
	\label{fig:real_data}
\end{figure*}

The results from the different approaches are visualized in \cref{fig:real_data}.
For FT and the replay-based approach, we observe the highest deterioration on the previous task.
In FT, there are no remedial measures to prevent forgetting and the previously learned information is overwritten.
The replay-based method is affected by the propagation of erroneously generated images and the corresponding pseudo labels.
ILT exhibits low-quality segmentation results on both tasks.

\subsection{Unseen Domains}
\label{sec:results:unseen}
To highlight the efficacy of our proposed approach, we evaluate it further on unseen datasets.
The emphasis here is not on the quantitative results achieved on these datasets but rather on the generalization of our approach to unseen data.
Dark Zurich and Nighttime Driving, both contain entirely nighttime images, our pipeline correctly identifies the adversarial domain with 100~\% accuracy.
Thus, all samples are directed to the ACDC expert achieving mIoUs of 50.56 and 55.84 respectively compared to mIoUs of 11.46 and 19.17 by the CS model.

\subsection{Fine-grained Adversarial Domains}
\label{sec:results:SHIFT}
SHIFT is a large dataset, allowing for a fine distinction between adverse driving conditions.
Our setup consists of one ideal clear daytime domain and four adverse domains of night, rain, fog, and overcast.
The results on SHIFT are presented in \cref{tab:cat2}.
The JT model leverages the large amount of data and improves the results across all domains.
Intuitively, the upper bound for PSS should be the corresponding single-model results.
However, we observe that in certain cases our approach improves over the corresponding baseline.
This is due to the routing by the task-experts, which determine the most suitable model regardless of the samples' original domain.
We can notice the invariance of PSS to the extent of the domain gap.
For large gaps, the classification works nearly perfectly. 
For small gaps where domains converge, the domain expert processing the sample becomes irrelevant.

\begin{table*}[t]
	\centering
	\caption{Results on SHIFT with a finer distinction between ideal and adverse driving conditions.
		Our proposed approach almost reaches the theoretical upper limit of joint training.}
	\resizebox{\linewidth}{!}{
		\begin{tabular}{c@{\hspace{0.35cm}}
				c@{\hspace{0.1cm}}c@{\hspace{0.35cm}}
				c@{\hspace{0.1cm}}c@{\hspace{0.35cm}}
				c@{\hspace{0.1cm}}c@{\hspace{0.35cm}}
				c@{\hspace{0.1cm}}c@{\hspace{0.35cm}}
				c@{\hspace{-0.1cm}}c@{\hspace{0.5cm}}
				c
			}
			
			\boldhline
			\textbf{Method} & \textbf{Day} &  & \textbf{Night} & & \textbf{Fog} & & \textbf{Rain} & & \textbf{Overcast} & & \textbf{Avg.} \\
			\boldhline
			
			Single Task & 83.56 &  & 77.77 &  & 78.53 &  & 84.10 &  & 83.57 & & -- \\
			
			\hline
			
			Fine-Tuning & 74.77 & \textcolor{red}{(-08.79)} & 29.48 & \textcolor{red}{(-48.29)} & 55.87 & \textcolor{red}{(-22.66)} & 72.62 & \textcolor{red}{(-11.48)} & 84.19 & \textcolor{blue}{(+00.62)} & 63.38 \\
			
			
			\textbf{PSS (Ours)} & 83.33 & \textcolor{red}{(-00.23)} & 77.77 & \textcolor{blue}{(+00.00)} & 79.42 & \textcolor{blue}{(+00.89)} & 83.44 & \textcolor{red}{(-00.66)} & 83.70 & \textcolor{blue}{(+00.13)} & \textbf{81.53} \\
			
			\hline
			
			Joint Training & 84.15 & \textcolor{blue}{(+00.59)} & 78.43 & \textcolor{blue}{(+00.66)} & 81.07 & \textcolor{blue}{(+02.54)} & 84.63 & \textcolor{blue}{(+00.53)} & 84.52 & \textcolor{blue}{(+00.95)} & 82.56 \\
			\boldhline
			
		\end{tabular}
	}
	\label{tab:cat2}
\end{table*}

\begin{figure*}[t]    
	\centering
	\includegraphics[width=\textwidth]{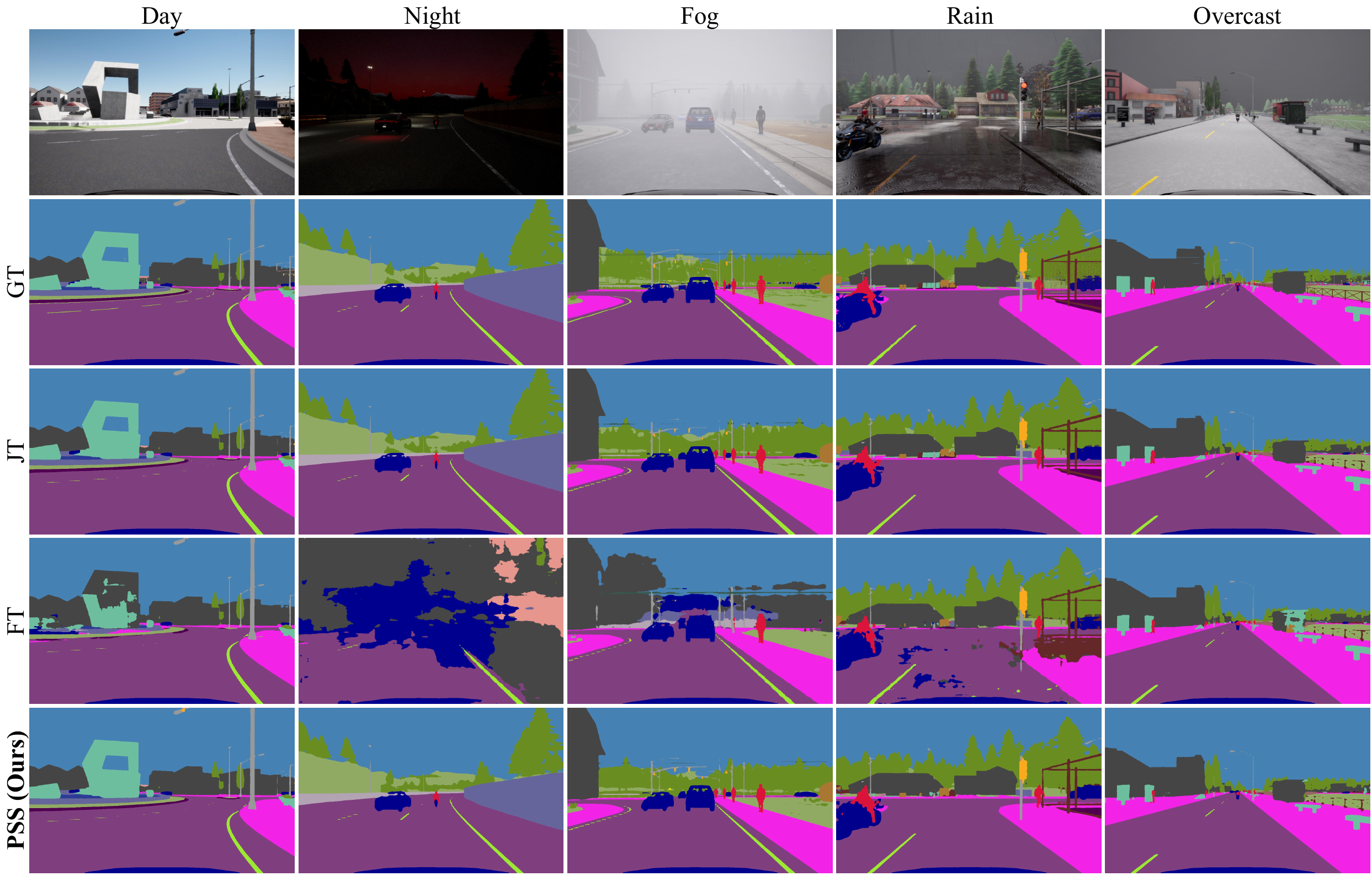}
	\caption{Qualitative visualization of segmentation masks on SHIFT \cite{SHIFT}. Each adversarial condition is considered individually resulting in a total of five domains. Our proposed Progressive Semantic Segmentation (PSS) achieves results that are qualitatively on par with the jointly trained model (JT) and very close to the ground truth (GT).}
	\label{fig:shift}
\end{figure*}

The results of the fine-grained adversarial domains from SHIFT are presented in \cref{fig:shift}.
For the fine-tuning approach (FT), we can observe the performance increasingly worsen along the sequence of tasks.
The information lost is the highest in the case of significantly different domains such as night, which introduces domain-specific characteristics for a few classes and challenging conditions.
Our PSS achieves results comparable to single-task domain experts and even improves for a few domains.

\subsection{Runtime Analysis}
\label{sec:results:time}
\begin{figure}[t]
	\includegraphics[width=\columnwidth]{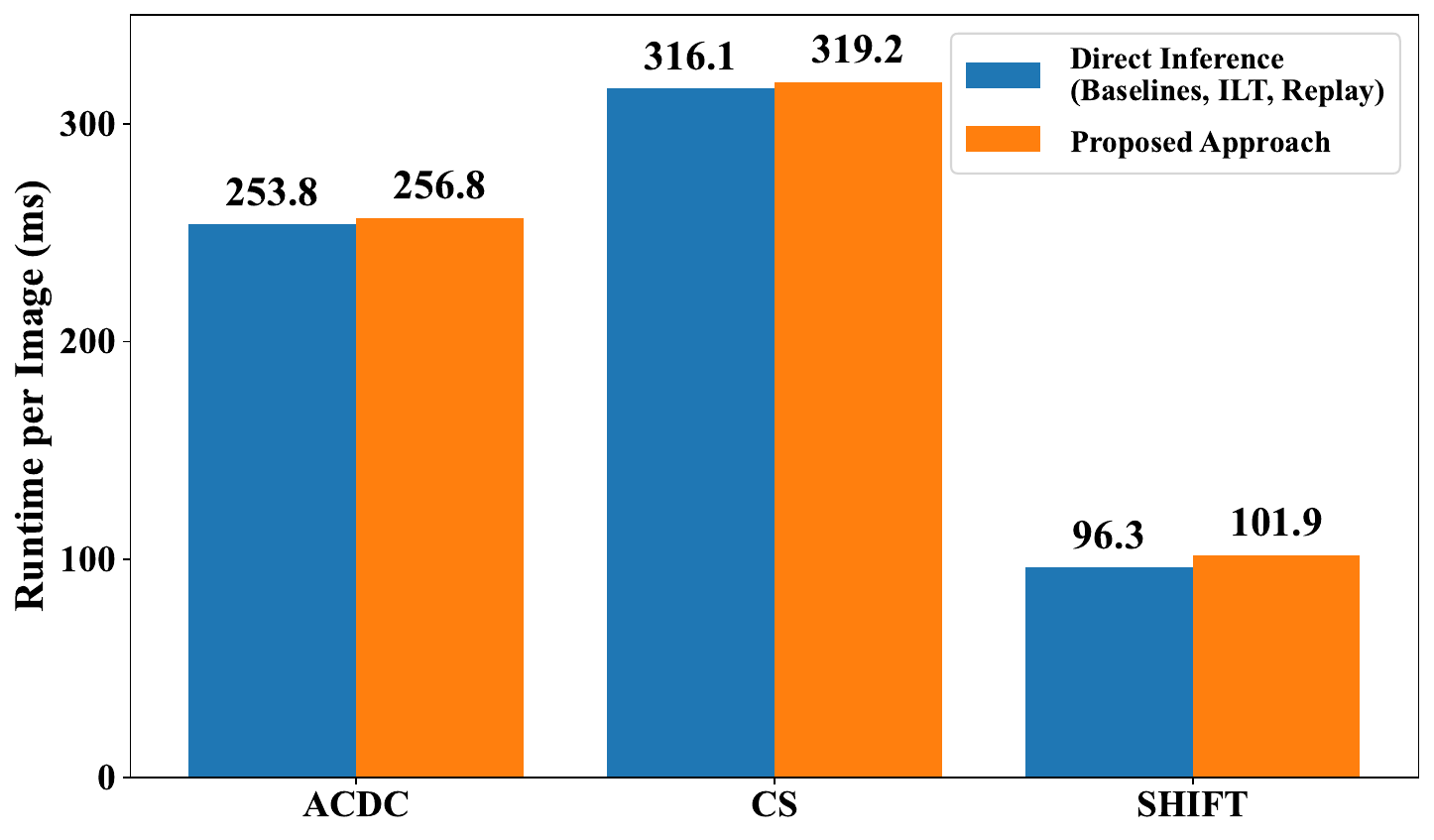}
	\caption{Computational overhead of PSS \vs direct inference. A minuscule increase in inference time of 3-6 ms is incurred for the reconstruction of the domain and routing to the domain expert. The direct inference refers to all other methods that do not involve domain inference prior to segmentation, including the baselines, as well as replay, and regularization-based methods.}
	\label{fig:inference_time}
\end{figure}

In our pipeline, prior to segmentation, the domain is inferred through reconstruction by the collection of AEs, and the corresponding domain expert is selected.
We acknowledge the overhead of this architecture-based incremental learning approach.
Therefore, we delve into the specifics of the computational costs for reconstructing the images and inferring the domain.
To evaluate our runtime performance, we compare it to \textit{direct inference}, which resembles the task-incremental learning in the single-task baseline where the task-ID is explicitly provided. \textit{Direct inference} also encompasses all other approaches that use a single model and do not require domain inference before segmentation. This category includes baseline methods such as joint training, fine-tuning, as well as techniques based on regularization \cite{ILT} and replay \cite{ganformer}.
We use a single NVIDIA A100 GPU for inference and report the average runtime for different datasets in \cref{fig:inference_time}.
We report that for the coarse distinction between CS and ACDC, the computational overhead is 3 ms, whereas the distinction within  SHIFT categories involves reconstruction by 5 task-experts and the overhead is 5.6 ms.

\subsection{Hybrid Incremental Learning}
\label{addExp:hybridIL}
Domain-incremental learning learns from domains with different input distributions under the constraint that the set of classes remains consistent.
In class-incremental learning, the input distribution remains the same and non-overlapping classes are added.
A non-incremental shift in the output space curtails the use of existing CL approaches and even the joint training becomes more challenging.
When new domains are added which may have overlapping classes, a conflict arises.
For instance, the previously seen vehicle class may have further distinction into cars, buses, and domain-specific classes such as auto-rickshaws.
Our proposed approach is devoid of these limitations and is able to handle both varying input and output distributions with different and overlapping classes. 

To demonstrate this hybrid incremental learning, we use CS and IDD.
The \textit{joint training} on both domains, requires a mapping of classes into a common label space.
We map the classes of IDD to the corresponding classes of CS, and ignore those that have no counterpart.
As a result, the \textit{joint model} is always evaluated on 19 classes only.
\Cref{table:hybridIL} presents the results.
Even in this hybrid setting with overlapping classes, our PSS achieves results close to the single-task baseline.
From this we infer that it may be beneficial to treat an unconstrained domain with diverse classes as an adversarial condition, necessitating a domain expert.

\begin{table}[t]
	\centering
	\caption{Results on the hybrid incremental learning approach with both varying input and output distributions. The joint model is trained and evaluated on the 19 classes from CS. The single-task model and our approach is evaluated on 19 classes for CS and 26 classes for IDD.} 
	\label{table:hybridIL}
	\begin{tabular}{cc@{\hspace{0.1cm}}cc@{\hspace{0.1cm}}c}
		\boldhline
		\textbf{Method} & \textbf{CS} & & \textbf{IDD} & \\
		\boldhline
		Single Task & 80.55 & & 72.91 \\
		\hline
		\textbf{PSS (Ours)} & 80.22 &\textcolor{red}{(-00.33)}  & 71.75 & \textcolor{red}{(-01.16)} \\
		\hline
		Joint Training & 81.53  & \textcolor{blue}{(+00.98)} & \hphantom{*}82.53* \\
		\boldhline
		\multicolumn{5}{r}{\footnotesize *Evaluated on 19 instead of 26 classes}
	\end{tabular}
\end{table}

\begin{table}[t]
	\centering
	\caption{Results in mAP for Progressive Object Detection (POD) on the day and nighttime conditions of SHIFT.
		Our approach can be directly integrated into any pipeline repurposing single-task models as domain experts.}
	
	\begin{tabular}{cc@{\hspace{0.1cm}}cc@{\hspace{0.1cm}}c}
		\boldhline
		\textbf{Method} & \textbf{Day} & & \textbf{Night} & \\
		\boldhline
		
		Single Task & 36.23 & & 33.77 & \\ 
		\hline
		Fine-Tuning & 28.76 &  \textcolor{red}{(-07.47)} & 35.29 & \textcolor{blue}{(+01.52)}\\
		\textbf{POD (Ours)} & 36.15 & \textcolor{red}{(-00.08)}  & 33.77  & \textcolor{blue}{(+00.00)} \\
		\hline
		Joint Training & 36.41 & \textcolor{blue}{(+00.18)} & 34.24 & \textcolor{blue}{(+00.47)} \\
		\boldhline
		
	\end{tabular}
	\label{table:objDet}
\end{table}

\subsection{Transfer to Object Detection}
\begin{figure}[t]    
	\centering
	\includegraphics[width=\columnwidth]{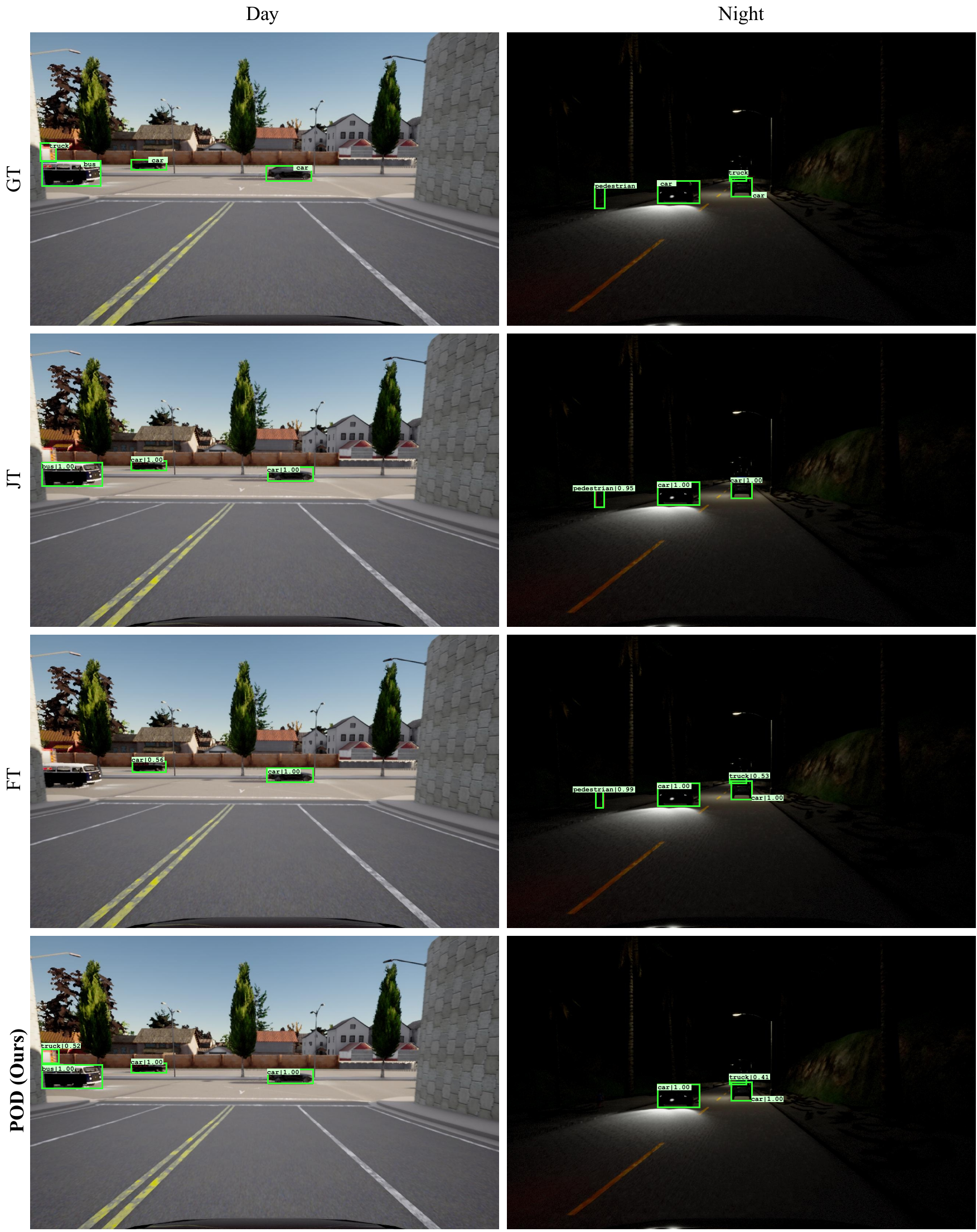}
	\caption{Domain-incremental learning for object detection by Progressive Object Detection (POD) on the SHIFT dataset \cite{SHIFT}. }
	\label{fig:pod}
\end{figure}
\label{sec:results:detection}
Our proposed approach of using AEs to dynamically identify the task during inference can be implemented across different computer vision tasks.
It resolves the limitation in architecture-based CL approaches which typically require identifying expert models \cite{progressiveNN} or dedicated heads \cite{MDIL} or domain-specific statistics \cite{DIL_DET} for routing of the test image.
In this experiment, we highlight this generalizability by applying our progressive approach to object detection, for which catastrophic forgetting is severe \cite{witte2023severity}.
For this, we consider two domains namely the day and nighttime conditions of SHIFT. 
The pipeline is similar to the one illustrated in \cref{fig:Overview}, with the exception that the segmentation networks $S_k$ are substituted by object detection models.
We use Faster-RCNN \cite{FasterRCNN} with ResNet-101 as the backbone and the results in mean average precision (mAP) are compared against the single-task, fine-tuned, and jointly trained models and presented in \cref{table:objDet}.
Through this, we would like to reiterate the versatility of our approach which can be directly integrated for incremental learning without the need for retraining with regularization or training of generative models.
The results are presented in \cref{fig:pod} and we once again observe the highest forgetting in fine-tuning (FT).
In the domain inference, the task experts achieve nearly 100\% accuracy in distinguishing between the two domains and our approach, Progressive Object Detection (POD), mitigates forgetting and produces reasonable results on the second domain.


\section{\uppercase{Limitations}} \label{sec:limitations}
A common criticism and limitation associated with architecture-based methods, where the number of models increases linearly with the number of tasks/domains, is scalability.
The individual models that are leveraged as domain experts in our work cannot be extended indefinitely for practical reasons.
However, we believe that for reasonable number of domains, architecture-based methods are feasible.
Similarly, the inference time increases linearly with the number of domains.
Our analysis in \cref{sec:results:time} shows that there must be hundreds of domains before the overhead reaches the time complexity of the segmentation model. 
At the same time, scalability is not only an issue with architecture-based methods.
Other approaches such as replay-based methods, may require training and maintaining a generative model for every task.
Lastly, our work is focused on the domain gap between varying weather and illumination conditions.
However, there are many other dimensions with respect to domain-specific environmental conditions.
Covering all possible aspects can result in a combinatorial explosion of domain experts.


\section{\uppercase{Conclusion}}
\label{section:conclusion}
Progressive Semantic Segmentation (PSS) addresses the problem of continuous adaptation to changing environments for autonomous driving systems from the perspective of continual learning. 
It employs a dynamically growing collection of domain experts, each of which is trained on an individual domain.
This approach mitigates forgetting to a great extent. 
To make PSS task-agnostic, we use a collection of task experts to dynamically infer the domain during inference.
Our experiments demonstrate superior performance in comparison to previous domain-incremental methods and highlight the flexibility of PSS in unseen domains, in hybrid incremental scenarios, and for other vision tasks like object detection.
In future work, we would like to combine PSS with domain adaptation techniques to better exploit the knowledge of previous models for new tasks.

\section*{ACKNOWLEDGEMENTS}
This work was partially funded by the Federal Ministry of Education and Research Germany under the projects DECODE (01IW21001) and COPPER (01IW24009).

\bibliographystyle{apalike}
{\small
\bibliography{main}}

%

\end{document}